\documentclass{article}

\usepackage[final]{neurips_2022}

\usepackage[utf8]{inputenc} 
\usepackage[T1]{fontenc}    
\usepackage{hyperref}       
\usepackage{url}            
\usepackage{booktabs}       
\usepackage{amsfonts}       
\usepackage{nicefrac}       
\usepackage{microtype}      
\usepackage{xcolor}         

\usepackage{float}

\usepackage{graphicx}
\usepackage{multirow}
\usepackage{amsmath,amsfonts,amssymb}
\usepackage{dsfont}
\usepackage{enumitem}
\usepackage{xcolor}
\usepackage{array}
\usepackage{makecell}
\usepackage[export]{adjustbox}
\usepackage{hyperref}
\usepackage{ifthen}
\usepackage{wrapfig}

\usepackage[capitalize]{cleveref}
\crefname{section}{Sec.}{Secs.}
\Crefname{section}{Section}{Sections}
\Crefname{table}{Table}{Tables}
\crefname{table}{Tab.}{Tabs.}

\newcommand{\encoder}[1]{$E$}
\newcommand{\decoder}[1]{$D$}
\newcommand{\maskoutputv}[1]{$\hat{x}_v$}
\newcommand{\maskoutputa}[1]{$\hat{x}_a$}

\newcommand{\imagenet}[1]{ImageNet}
\newcommand{\Yttemp}[1]{YTTemporal180M}
\newcommand{\Yttsubset}[1]{YTT-S}
\newcommand{\howtohundredM}[1]{HowTo100M}
\newcommand{\vqa}[1]{VQAv2}
\newcommand{\vqaold}[1]{VQAv1}
\newcommand{\vqaonek}[1]{VQA-1k}
\newcommand{\msrvtt}[1]{MSR-VTT}
\newcommand{\youcook}[1]{Youcook2}
\newcommand{\crosstask}[1]{CrossTask}
\newcommand{\nlvr}[1]{NLVR2}
\newcommand{\cmumosei}[1]{CMU-MOSEI}
\newcommand{\flickr}[1]{Flickr30K}
\newcommand{\places}[1]{Places-205}
\newcommand{\placesaudio}[1]{Places-400k}

\newcommand{\methodname}[1]{\textsc{TVLT}}
\newcommand{\avlnet}[1]{AVLnet}

\newcommand{\VL}[1]{Vision-and-Language}
\newcommand{\vl}[1]{vision-and-language}

\newcolumntype{P}[1]{>{\centering\arraybackslash}p{#1}}

\setcitestyle{square,numbers}
\hypersetup{colorlinks,linkcolor={blue},citecolor={green},urlcolor={magenta}}

\title{
{TVLT: Textless Vision-Language Transformer}
}

\newcommand*\samethanks[1][\value{footnote}]{\footnotemark[#1]}

\author{
Zineng Tang\thanks{equal contribution} \quad Jaemin Cho\samethanks \quad Yixin Nie\samethanks \quad Mohit Bansal \\
UNC Chapel Hill\\
\{\tt\small terran, jmincho, yixin1, mbansal\}@cs.unc.edu
}

\begin{document}

\maketitle

\setcounter{footnote}{0}

\begin{abstract}

In this work, we present the Textless Vision-Language Transformer (TVLT), where homogeneous transformer blocks take raw visual and audio inputs for vision-and-language representation learning with minimal modality-specific design, and do not use text-specific modules such as tokenization or automatic speech recognition (ASR).
TVLT is trained by reconstructing masked patches of continuous video frames and audio spectrograms (masked autoencoding) and
contrastive modeling to align video and audio.
TVLT attains performance comparable to its text-based counterpart on various multimodal tasks,
such as
visual question answering,
image retrieval, video retrieval,
and multimodal sentiment analysis,
with 28x faster inference speed and only 1/3 of the parameters.
Our findings suggest the possibility of learning compact and efficient visual-linguistic representations from low-level visual and audio signals without assuming the prior existence of text.\footnote{Our code and checkpoints are available at: \url{https://github.com/zinengtang/TVLT}}

\end{abstract}

\section{Introduction}

Humans perceive and learn the external world through signals from multiple modalities.
To embody such human learning in machines, substantial research efforts are dedicated to developing vision-and-language (VL) models that can understand the joint semantics between visual and linguistic modalities and solve tasks such as visual question answering~\cite{antol2015vqa}.
Although most such VL models use written language rather than spoken language as the main verbal communication channel, the default communication modality among humans has been speech, since circa 100,000 BCE~\cite{tattersall2009cognitive}.
Written language is relatively recent; cuneiform script, the earliest writing system, was developed circa 3,200 BCE~\cite{schmandt2014evolution}.
Moreover, we have witnessed an increasing usage of AI models in real-world products such as virtual assistants and smart speakers~\cite{10.1093/jamiaopen/ooz011}, where perception-level signals such as video and audio are the natural form of input. Intuitively, direct modeling of such signals will potentially yield more compact and efficient representations.

\begin{figure*}[t]
  \centering
\includegraphics[width=.95\textwidth]{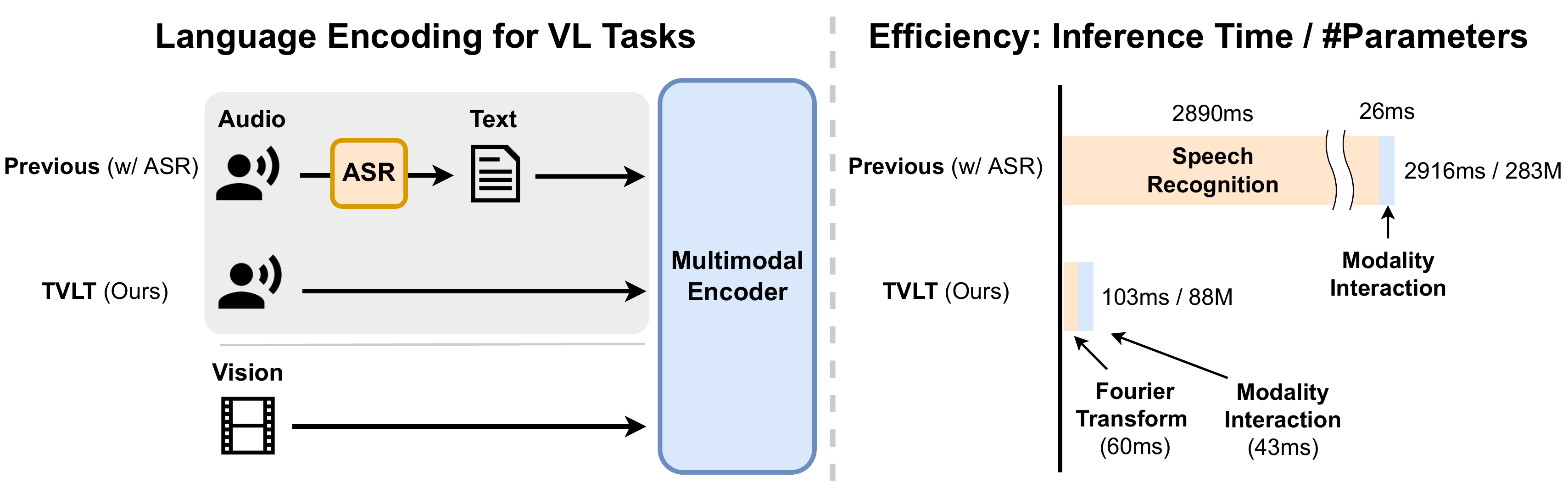}
  \caption{Comparison of previous VL architectures and our proposed textless framework \methodname{}.
  The removal of automatic speech recognition (ASR) from the VL pipeline brings efficiency improvement while maintains competitive performance.
  For inference time calculation, we use 8 video frames and 20s audio (see \cref{sec:efficiency} for detail). As shown in \Cref{tab:video_ablation}, \methodname{} achieves competitive performance to text-counterpart on video retrieval and multimodal sentiment analysis tasks.
}
\label{fig:teaser}
\end{figure*}

Transformers~\cite{vaswani2017attention} have recently achieved great success in vision-language representation learning~\cite{tan2019lxmert,chen2019uniter,lu2019vilbert,sun2019videobert,zellers2021merlot,Zellers2022MerlotReserve} by using text-based modules~\cite{devlin2018bert}
on text-annotated images or videos.
However, 
it is non-trivial to learn VL representations using transformers that take 
only low-level visual and acoustic inputs without the prior existence of written language.
The challenge lies in the difference between text and acoustic signals; text is discrete and dense in information, while acoustic signals are continuous and sparse in information~\cite{he2021masked,baade2022mae-ast}. 
Therefore, modality-specific architectures have been used to model data from different modalities.
It is only recently that researchers started using modality-agnostic transformer architecture to learn representations of different unimodal~\cite{dosovitskiy2020image,gong2021ast,Baevski2022Data2Vec}, vision-text~\cite{Kim2021ViLT,Nie2022MLPVL}, or vision-audio-text~\cite{akbari2021vatt} data.
However, to the best of our knowledge, no previous work has explored a single homogeneous (modality-agnostic) minimalist transformer that learns visual-linguistic representations directly from visual and acoustic input at the perception level (without relying on text),
and also makes the textless VL model more compact and efficient than the existing text-based VL models (see \cref{sec:related_work} for details).

In this work, we propose Textless Vision-Language Transformer (\methodname{}) for \vl{} representation learning based on video data as the natural source of raw visual and audio input.
As depicted in \cref{fig:architecture},
\methodname{} accepts low-level video frames and audio spectrograms as input.
We employ a minimalist design for \methodname{} where homogeneous transformer blocks are used for both the encoder and decoder.
\methodname{} is trained by reconstructing masked patches of continuous video frames and audio spectrograms (masked autoencoding) and
contrastive modeling to align video and audio.
More importantly, \methodname{} makes no assumptions about the existence of written language and does not involve explicit modeling of text input, such as automatic speech recognition (ASR) or tokenization, which are crucial submodules in the success of existing VL models in aligning written concepts with visual clues.

Despite the removal of text-based modules and modality-specific designs, \methodname{} achieves results comparable to its text-based counterparts in multimodal tasks
(with either direct audio input, or text converted to audio input via TTS)
such as
visual question answering,
image retrieval, video retrieval,
and multimodal sentiment analysis,
while being computationally efficient with 1/3 parameters and a 28x faster inference speed, as illustrated in \cref{fig:teaser}.
This indicates that the removal of text-specific modules such as ASR in \vl{} modeling helps reduce computational redundancy in existing pipelined learning paradigms, where text is first extracted through ASR and then further processed by a text-based VL model.
Furthermore, we also show that \methodname{} can capture acoustic information beyond speech and is more effective in multimodal emotion classification than its text-based counterpart.
We hope that our findings spark further research in the realm of textless VL models that take raw signals as input and seek to learn a more compact and efficient \vl{} representation.
\section{Related Work}
\label{sec:related_work}

\paragraph{Text-based Representation Learning.}
\looseness=-1 Large-scale unsupervised pretraining of contextualized language models based on written texts has seen great success in recent years. ELMo~\cite{peters2018deep} proposes to pretrain and finetune a large recurrent language model, which improves performance on a diverse set of downstream natural language processing tasks.
BERT~\cite{devlin2018bert} improves the scalability of the pretrain-then-finetune paradigm by using a transformer~\cite{vaswani2017attention} model with a masked language modeling objective.
Since then, the pre-training of transformers has been extensively explored for transfer learning in language~\cite{liu2019roberta,yang2019xlnet,lan2019albert,dong2019unified,song2019mass,raffel2020exploring,clark2020electra}. In these methods, learning is focused on eliciting high-level linguistic semantics and structures from unlabeled written texts or natural sequences of words.

\paragraph{Audio-based Representation Learning.}
Pretraining methods on audio input involve transferring the continuous 1D audio signal into dense vectors that can be input to a speech or acoustic model. Early work mainly uses recurrent neural networks~\cite{chung2016audio, chung2018speech2vec, shenoy-sardana-2020-multilogue} and convolution networks~\cite{schneider2019wav2vec} for audio encoding.
To take advantage of the proven expressiveness and genericity of transformers, more recent work proposed using audio spectrograms~\cite{gong2021ast, gong2021ssast, baade2022mae-ast} as image input and then encoding the patches of such images with a transformer, following the same methodology in computer vision~\cite{dosovitskiy2020image}. The pretraining objectives for transformers range from classification~\cite{gong2021ast} to masked audio modeling ~\cite{gong2021ssast, baade2022mae-ast}.
A line of work uses an audio transformer with discrete audio units for pretraining~\cite{Hsu2021HuBERTSS} and speech tasks such as generative spoken language modeling \cite{lakhotia-etal-2021-generative, kharitonov-etal-2022-text} and
speech emotion conversion \cite{Kreuk2021TextlessSE}.
These works focus on learning the acoustic and linguistic characteristics of a language from raw audio or spectrogram.

\paragraph{Vision-and-Language Representation Learning.}

Following the success of pretraining of transformer language models, pretraining of
image+text~\cite{tan2019lxmert,lu2019vilbert,chen2019uniter,li2020unimo,zhou2020unified,li2020unicoder},
video+text~\cite{sun2019videobert,miech2019howto100m,zhu2020actbert,miech2020end,li2020hero,tang2021decembert,zellers2021merlot},
and video+text+audio~\cite{Tsai2019MulT,zadeh2018multimodal,Rahman2020MAGBERT,Zellers2022MerlotReserve,akbari2021vatt}
multimodal transformers has recently achieved improvements in downstream VL tasks such as visual question answering~\cite{antol2015vqa,Hudson2019} and text-to-video retrieval~\cite{xu2016msr,zhou2018towards}.
These methods use text, such as written captions or ASR transcripts, as input into the language channel.
There is another line of work on models taking video+audio input, where they can utilize naturally synchronized vision+audio pairs from videos.
Audio-visual synchronization is often used for self-supervised learning~\cite{av_eccv16_abSound,av_iccv17_look,av_eccv18_Owens,av_nips18_coop,av_nips20_CrossLabelling,av_cvpr21_RAVID,av_iclr21_activeContrastive},
or for downstream tasks such as automatic speech recognition~\cite{Afouras2018,Shillingford2019,shi2022avhubert}
and video retrieval~\cite{Surs2018CrossmodalEF,rouditchenko2020avlnet,Sanabria2021MILAN,ECLIPSE_ECCV22}.
Our work is different from these works, in that
we focus on the design of a homogeneous and modality-agnostic transformer~(\cref{sec:method}) to achieve a novel, unified, and minimalist textless visual-linguistic representation learning method directly from visual and acoustic signals (without relying on text), via masked autoencoding and contrastive modeling objectives~(\cref{sec:pretrain_objectives}),
which also makes the textless VL model more compact and efficient than the existing text-based VL models.
\section{\methodname{}: \textbf{T}extless \textbf{V}ision-\textbf{L}anguage \textbf{T}ransformer}
\label{sec:method}

\begin{figure*}[t]
  \centering
  \includegraphics[width=0.95\textwidth]{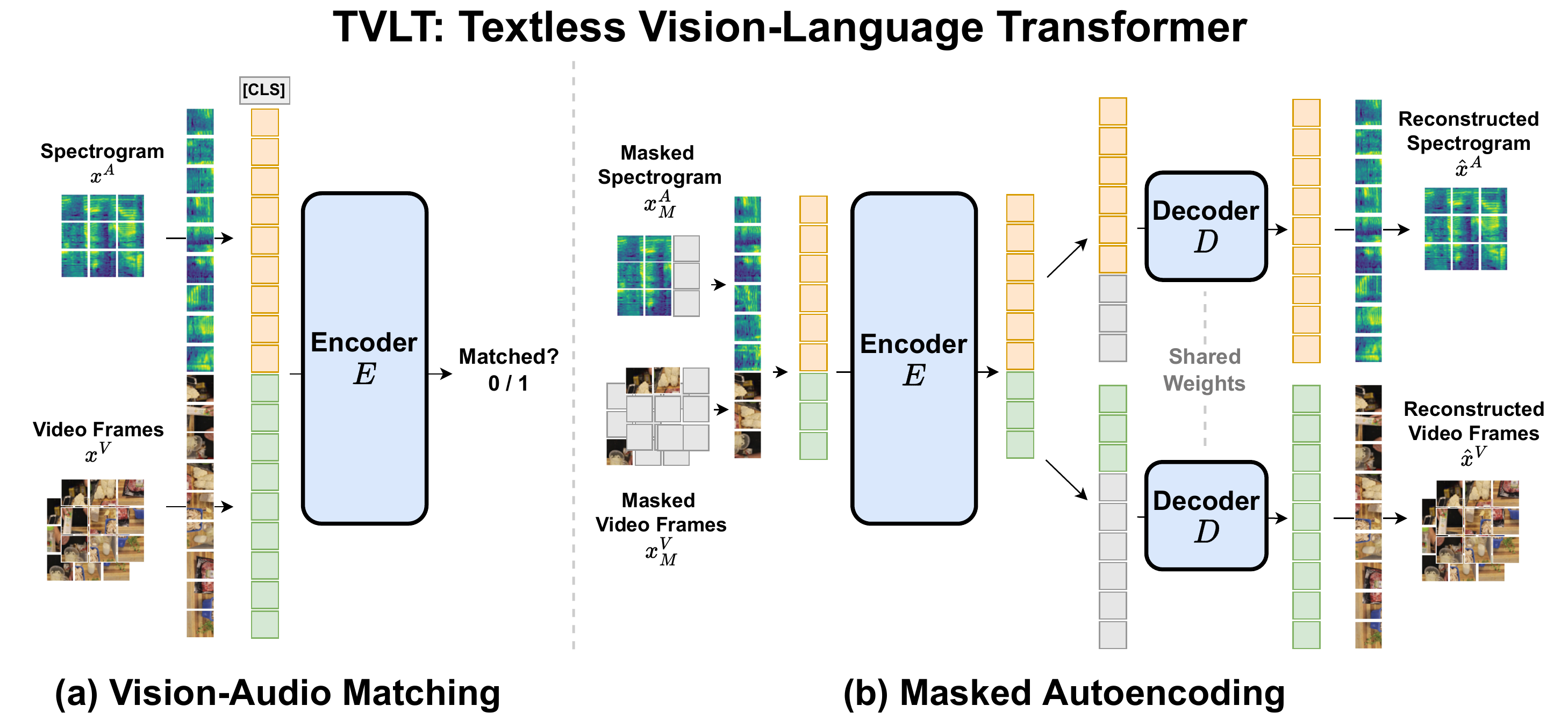}
  \caption{
    \methodname{} is pretrained with two objectives: (a) vision-audio matching (\cref{sec:vision_audio_matching}) and (b) masked autoencoding (\cref{sec:masked_autoencoding}).
    The model takes video frames and audio spectrogram as inputs and does not use text input and completely removes text from the pipeline.
}
\label{fig:architecture}
\end{figure*}

We introduce \methodname{}: \textbf{T}extless \textbf{V}ision-\textbf{L}anguage \textbf{T}ransformer, a minimal end-to-end \vl{} transformer model that accepts a list of embeddings obtained directly from perception-level video and audio input \textit{without text-specific modules}, as depicted in \cref{fig:teaser} and \cref{fig:architecture}.

\subsection{Input Embeddings}
\label{sec:embeddings}

The input embeddings of \methodname{} are the sum of
(1) modality embedding,
(2) temporal/spatial embedding for video,
(3) temporal/frequency embedding for audio, and
(4) vision/audio patch embedding.
As illustrated by the red and blue boxes in~\cref{fig:architecture}, the modality embeddings are two trainable vectors added to the input embeddings and used to indicate whether the input is from vision or audio input.
In what follows, we explain the details of vision and audio embeddings.

\paragraph{Vision Embeddings.}
We adopt ViT~\cite{dosovitskiy2020image}-style vision embedding, where each video frame of $224\times 224$ pixels is divided into a list of $16\times 16$-sized patches. Then, a liner projection layer is applied to the normalized pixel values of each patch, resulting in a 768-dimensional patch embedding.
For a video clip with N frame samples, the input tensor with shape $N\times 224\times 224 \times 3$ (time $\times$ height $\times$ width $\times$ channel) will result in $N\times 14\times 14$ embeddings.
The temporal and spatial embeddings are different trainable vectors added to the time, height, and width axis of the $N\times 14\times 14$ embeddings to incorporate the temporal and spatial information for each input patch.
We treat image input as a single frame video so that our model can handle both image and video tasks without modification of the architecture~\cite{bain2021frozen}. Temporal embedding is only added for video inputs; we do not use temporal embedding for images.

\paragraph{Audio Embeddings.}
To obtain audio embeddings, we first convert the 1D waveform of the raw audio signal to 128-dimensional log Mel-spectrogram having a dimension of $T\times 128$ (time axis $\times$ frequency axis).\footnote{
We use \texttt{melspectrogram} method of \texttt{librosa}~\cite{brian_mcfee_2022_6097378} with arguments: \texttt{sampling rate=44100, n\_fft=2048, hop length=512, window=`hann', pad\_mode=`constant', n\_mels=128}.}
Then, we treat the audio spectrogram as an image, divide the spectrogram images into patches, and apply a liner projection layer on each patch to obtain a 768-dimensional patch embedding.
This follows the audio embedding methods in recent work ~\cite{gong2021ast,gong2021ssast,baade2022mae-ast}, where a similar modality-agnostic transformer is used to model spectrogram patches. We experiment with two different patch sizes: $16\times 16$ (square patches similar to the vision modality) and $2\times 128$ (the same area as the first one but covers the entire frequency domain with a shorter time range) and use trainable temporal and frequency embeddings to indicate the temporal and frequency information of patches.\footnote{With 16x16 patch, a 20-second audio will have a spectrogram with shape $640\times 128$ (time axis $\times$ frequency axis), resulting in $40\times 8=320$ patches.}

\subsection{Multimodal Encoder-Decoder}
\label{sec:enc_dec}

The main architecture of \methodname{} is a transformer \cite{vaswani2017attention} consisting of a 12-layer encoder (hidden size 768), \encoder{}, and an 8-layer decoder (hidden size 512), \decoder{}.
We follow \citet{he2021masked} and use a shallow decoder that only serves for masked autoencoding objective~(\cref{sec:masked_autoencoding}) and has much fewer computations than the encoder.
After pretraining, we only use the encoder representation for finetuning on downstream tasks.

\section{Pretraining Objectives}
\label{sec:pretrain_objectives}

By virtue of our minimal and modality-agnostic design, \methodname{} is pretrained with two objectives:
(1) vision-audio matching (\cref{sec:vision_audio_matching}) and (2) masked autoencoding (\cref{sec:masked_autoencoding}).
For each training batch,
we compute each objective through a separate forward pass and use the weighted sum of them for the final loss, where $\lambda^{\text{VAM}}=1.0$ and $\lambda^{\text{MAE}}=0.3$.
\begin{equation}
    loss = \lambda^{\text{VAM}}loss^{\text{VAM}} + \lambda^{\text{MAE}}loss^{\text{MAE}}
    \label{eq:loss}
\end{equation}

\subsection{Vision-Audio Matching}
\label{sec:vision_audio_matching}

We use the vision-audio matching (VAM) objective to learn the global cross-modal representation, as illustrated in \cref{fig:architecture} (a).
For each video input, we create a (positive) vision-audio pair $(x^{V+}, x^A)$.
Then, we construct half of the vision-audio pairs inside a batch as mismatched (negative) pairs $(x^{V-}, x^A)$, by replacing video frames $x^{V+}$ with randomly sampled video frames $x^{V-}$ from the training dataset.

Following previous \vl{} transformers~\cite{tan2019lxmert,chen2019uniter,lu2019vilbert,Kim2021ViLT}, a linear layer with sigmoid activation is used as the classification head applied to the encoder output of the first \texttt{[CLS]} token to obtain the matching probability $p$.
Then we compute the binary cross-entropy loss as:
\begin{equation}
    loss^{\text{VAM}} = - y \log p
\end{equation}
where $y$ is 1 when the input vision-audio pair $(x^{V}, x^A)$ is matched and 0 otherwise.

\subsection{Masked Autoencoding}
\label{sec:masked_autoencoding}

In addition to the VAM objective to learn cross-modal representation,
we also use the masked autoencoding (MAE) objective to improve unimodal representations in the \vl{} settings, by masking random patches of visual frames and the audio spectrogram, and reconstruct missing inputs as shown in \cref{fig:architecture} (b).
Concretely, we randomly drop a portion of
visual $x^V$ and audio embeddings $x^A$,
then feed the remaining patch embeddings to the encoder \encoder{}.
We create inputs for the decoder \decoder{} by adding the dropped embeddings as trainable vectors \texttt{[MASK]} to the same location as the original input (gray boxes in \cref{fig:architecture} (b)). We also add the corresponding temporal, positional, and frequency embeddings to the decoder input. Note that the temporal, positional, and frequency embeddings of the encoder and decoder are separately parameterized.
We calculate the mean squared error between the reconstructed and original video frames and spectrograms:
\begin{align}
    loss^{\text{MAE}} = \frac{1}{N^V_M}\sum_{i \in masked} || x^V_i - \hat{x}^V_i||_2^2 + \frac{1}{N^A_M}\sum_{j \in masked} || x^A_j - \hat{x}^A_j ||_2^2 
\end{align}
where $N^V_M$ and $N^A_M$ are the number of masked patches for vision and audio, respectively.
We compute the loss only on masked patches, similar to BERT~\cite{devlin2018bert}.

To save computation, we slice the audio and video parts of the encoder output and feed them separately to the decoder,
rather than decoding the video frames and the audio spectrogram jointly. In \cref{sec:ablation}, we show that separate decoding achieves better finetuning performance, as well as better efficiency than joint decoding.

\subsection{Masking Strategy}
\label{sec:masking_strategy}

\paragraph{Vision Masking.}
Following MAE~\cite{he2021masked}, we randomly mask 75\% of the visual patches, and the masking is applied for each video frame independently.

\paragraph{Audio Masking.}
Following MAE-AST~\cite{baade2022mae-ast}, we randomly mask 75\% of the spectrogram patches. 
To better capture speech-related audio representation, we emphasize audio masking on speech audios.
We use Audiotok~\cite{Sehili_auditok_2021}, an audio activity detection tool, to determine speech spans based on the detection of events in the energy of the audio signal. Then, we apply the masking only on those audio spans. We use a probability of 15\%.
We include the details of speech span detection in appendix.

\section{Experimental Setup}
\label{sec:expsetup}

To compare the audio-based and text-based language representations for \vl{} tasks, we pretrain our TVLT and its text-based counterpart on video datasets. Then, we finetune the models on a set of downstream \vl{} datasets for evaluation.

\subsection{Text-based TVLT Counterpart}
\label{sec:text_baseline}
Our text-based TVLT counterpart has the same architecture as the vanilla TVLT with minor changes to accommodate text-based inputs. Firstly, we use sentence-piece~\cite{kudo2018sentencepiece} tokenizer and then map each token to trainable vectors to encode the raw text into embeddings, instead of converting the continuous input of frames or spectrograms into patch embeddings as in vanilla TVLT. Secondly, we follow the norm in mask language modeling~\cite{devlin2018bert} to use an affine layer as the decoder to recover masked words and set the mask ratio on text to be 15\%, instead of using a transformer decoder to reconstruct 75\% of the masked video and audio embeddings in vanilla TVLT.

\subsection{Pretraining Datasets}
\label{sec:pretraining_datasets}

\paragraph{\howtohundredM{}.}
We used \howtohundredM{}~\cite{miech2019howto100m}, a dataset containing 136M video clips of a total of 134,472 hours from 1.22M YouTube videos to pretrain our model. Our vanilla TVLT is pretrained directly using the frame and audio stream of the video clips. Our text-based TVLT is trained using the frame and caption stream of the video. The captions are automatically generated ASR provided in the dataset.
We used 0.92M videos for pretraining, as some links to the videos were invalid to download.

\paragraph{\Yttemp{}.} 
\Yttemp{}~\cite{zellers2021merlot} includes 180M video segments from 6M YouTube videos that spans multiple domains, and topics, including instructional videos from \howtohundredM{}~\cite{miech2019howto100m}, lifestyle vlogs of everyday events from the VLOG dataset~\cite{ignat2019identifying}, and YouTube’s auto-suggested videos for popular topics like `science' or `home improvement'.
Each video segment consists of 1) an image frame extracted from the middle timestep of the segment, and 2) an ASR-based caption of L=32 BPE~\cite{gage1994new, sennrich-etal-2016-neural} tokens. For each sample, we randomly sample a 15s video clip from the entire video to form a setting similar to \howtohundredM{} dataset. Concretely, the original dataset provides 100 label files which are random split of the dataset.
We sample 20\% of \Yttemp{} (0.93M videos) so that the resulting subset consists of a similar number of videos to \howtohundredM{} (0.92M videos), and call it \Yttsubset{}. In appendix, we show that pretraining \methodname{} on \Yttsubset{} can improve the downstream task performance of over pretraining on \howtohundredM{}.

\subsection{Downstream Tasks}
\label{sec:downstream_tasks}

We evaluate models on video-based and image-based \vl{} tasks to compare the learned representation based on audio and text.
For video-based tasks, we experiment with video retrieval~\cite{xu2016msr,zhou2018towards,zhukov2019cross} and multimodal sentiment analysis~\cite{zadeh2018multimodal}.
For image-based tasks, we experiment with image retrieval~\cite{young2014image} and visual question answering~\cite{antol2015vqa,goyal2017making}.
Although audio comes naturally with video, image-based tasks, such as visual question answering, do not include audio. Thus, we obtain audio queries for visual question answering via the text-to-speech (TTS) synthesis method (\cref{sec:TTS}).

\paragraph{Audio-to-Video Retrieval.}
Following \avlnet{}~\cite{rouditchenko2020avlnet}, we use \msrvtt{}~\cite{xu2016msr}, \youcook{}~\cite{zhou2018towards}, and \crosstask{}~\cite{zhukov2019cross} for audio-to-video retrieval. We also follow the same data split in \avlnet{}~\cite{rouditchenko2020avlnet} to finetune our models on their respective training set.

\msrvtt{} is an open domain video dataset, consisting of 10,000 video clips from 20 categories such as music, movies or food.
We follow \avlnet{} for the standard split, i.e., 6,783 training clips and 1000 test clips (where 32 videos do not have sound). We report the test split results.

\youcook{} is a video dataset on cooking tutorials that contains 2,000 long videos of 89 cooking recipes. Each recipe has on average 22 videos. It has 9,586 training clips and 3,350 validation clips. We report the validation split results.

\crosstask{} dataset contains instructional videos for 83 different tasks, divided into 18 primary tasks and 65 related tasks. Primary tasks are manually collected with temporal step human annotations and are the main focus of tasks such as cooking or repairing. Related tasks are automatically collected without any annotations and are tasks related to the primary tasks, such as masking latte (primary) vs. making machiato (related). The goal of related tasks is to assess whether they can improve primary tasks. 
It has 17,840 training clips and 2,819 validation clips. We report the validation split results. For all three tasks, we extract \texttt{mp3} audio from videos with a sample rate of 44.1kHz. We also used the extracted audio or its corresponding ASR as retrieval queries for our experiment.

\paragraph{Multimodal Sentiment / Emotion Analysis.}
We use \cmumosei{}~\cite{zadeh2018multimodal} for multimodal sentiment analysis.
The dataset is made up of 23,454 movie review clips with more than 65.9 hours of YouTube video by 1000 speakers that cover 250 distinct topics. Each video clip also comes with a ground-truth transcription written by the author of the video.
Following previous studies, we use the 15,288/4,830 train-test split and report the binary accuracy (A2) for sentiment analysis and weighted accuracy (WA) and F1 score on emotion classification over 6 emotion categories.

\paragraph{Audio-to-Image Retrieval.}
We use \placesaudio{} (The Places Audio Caption 400K Corpus) \cite{harwath2016unsupervised, harwath2017learning, harwath2018jointly} for audio-to-image retrieval.
The dataset contains approximately 1,000 hours of 400,000 spoken English captions for natural images drawn from the \places{} \cite{zhou2014learning} image dataset. The queries are conceptual descriptions of the image. The dataset also provides ASR of these audios. \places{} is a large-scale scene dataset with 205 scene categories such as forest, bedroom, and coast, which contains 2,500,000 images in total.

\paragraph{Visual Question Answering.}
We use \vqaold{}~\cite{antol2015vqa} and \vqa{}~\cite{goyal2017making} for visual question answering.
\vqaold{} contains 204,721 images from COCO~\cite{lin2014microsoft} and 430,725 questions.
\vqa{} is a newer version of \vqaold{}, with 265,016 images from COCO and 1,105,904 questions.
For experiments with audio questions,
we generate speech audio from textual questions using TTS (\cref{sec:TTS}) and report test-dev results for both tasks.

\paragraph{Finetuning on Downstream tasks.}
For each of the downstream tasks, we add a task-specific head (two-layer MLP) on top of the encoder representation.
For retrieval tasks, we use an MLP to map encoder representation of \texttt{[CLS]} to matching scores $\in [0,1]$,
which correspond to match vs. mismatch pairs,
and train the model jointly with binary cross-entropy loss.
For visual question answering tasks, we use an MLP to map the encoder representation of \texttt{[CLS]} to the answer probabilities with 3129 answer candidates, and train the model jointly with binary cross-entropy loss in a multi-label classification setup.
For multimodal sentiment analysis tasks, we use an MLP to map the encoder representation of \texttt{[CLS]} token to the entiment scores, and train the model jointly with L2 regression loss.

\subsection{Other Details}
\label{sec:training_details}

\paragraph{Automatic Speech Recognition (ASR).}
\label{sec:ASR}

For the text-based model mentioned above, we obtain text from audio with different automatic speech recognition (ASR) models.
We use the \texttt{asr-crdnn-rnnlm-librispeech} ASR model from the Speechbrain package~\cite{speechbrain}.
The model is based on RNN language model and CRDNN encoder-CTC/Attention decoder architecture and is trained on LibriSpeech \cite{panayotov2015librispeech}.
We also experiment with the Google Cloud Speech-to-Text API which uses Conformer~\cite{gulati2020conformer} as the backend model.\footnote{\url{https://cloud.google.com/speech-to-text}}

\paragraph{Text-to-Speech (TTS).}
\label{sec:TTS}

We use \texttt{WaveNet}~\cite{van2016wavenet} Google Cloud Text-to-Speech API\footnote{\url{https://cloud.google.com/text-to-speech/docs/wavenet}} to generate audio input for the questions in \vqa{}.
Since \vqa{} questions are written in English, we use a \texttt{en-US-neutral} speaker. We follow the default pitch and speech configuration.
We use the \texttt{mp3} audio format with a sample rate of 44.1kHz to match the audio configuration used in the pretraining.

\looseness=-1 \paragraph{Pretraining.} We train \methodname{} and the text-based TVLT counterpart
for 200k steps using Adam optimizer \cite{kingma2014adam} with a learning rate of 1e-5, batch size 4096, and a decay rate of 0.001 with a cosine schedule \cite{loshchilov2017decoupled}.
We initialize the weights of both models with the masked autoencoder transformer in \citet{he2021masked} that is pretrained on \imagenet{}~\cite{deng2009imagenet}.
For the pretraining objectives in \cref{eq:loss}, we use $\lambda^{\text{VAM}}=1.0$ and $\lambda^{\text{MAE}}=0.3$. 
For each video clip, we uniformly sample 8 frames. 
Pretraining takes 2 weeks with 4 NVIDIA RTX A6000 GPUs (each 49GB memory).

\paragraph{Finetuning on Downstream Tasks.}
We use a learning rate of 1e-5, batch size 256, and a decay rate of 0.001 with a cosine schedule for all tasks. For each video clip, we uniformly sample 8 frames. We use 2 NVIDIA RTX A6000 GPUs.

\section{Results and Analysis}
\label{sec:results}

\begin{table}[t]
\caption{
Comparison of \methodname{} and its text-based counterpart on audio-to-video retrieval and video-based multimodal sentiment analysis benchmarks;
\textit{HT100M}=\howtohundredM{},
\textit{YTT-S}=\Yttemp{} subset.}
\label{tab:video_ablation}
\centering
\resizebox{.9\textwidth}{!}{
\begin{tabular}{l ccc c ccc c c c}
\toprule
\multirow{2}{*}{Method} & \multicolumn{3}{c}{Input Mod.} & \multirow{2}{.1\textwidth}{\centering Pretrain Datasets} & \multicolumn{3}{c}{Audio-to-Video Retrieval (R@1) $\uparrow$} & Sentiment (A2) $\uparrow$  & Latency $\downarrow$ \\
\cmidrule(lr){2-4} \cmidrule(lr){6-8} \cmidrule(lr){9-9} \cmidrule(lr){10-10}
& V & T & A & & \msrvtt{} & \youcook{} & \crosstask{} & \cmumosei{} & (ms) \\
\midrule
\methodname{}  & \checkmark & \checkmark & & - & 3.1 & 5.0 & 2.2 & 68.1 & 2916\\
\methodname{}  & \checkmark & & \checkmark & - & 4.3 & 4.7 & 2.7 & 65.7 & 103\\
\methodname{}  & \checkmark & \checkmark & & HT100M & 17.1 & 24.9 & 11.1 & 76.5 & 2916\\
\methodname{}  & \checkmark & & \checkmark & HT100M & 22.6 & 31.8 & 14.9 & 75.3 & 103\\
\methodname{}  & \checkmark & \checkmark & & YTT-S & 19.3 & 26.3 & 12.2 & 76.6 & 2916\\
\methodname{}  & \checkmark & & \checkmark & YTT-S & \textbf{23.8} & \textbf{32.8} & \textbf{15.3} & \textbf{76.8} & 103\\
\bottomrule
\end{tabular}
}
\end{table}
\vspace{-5pt}

\begin{table}[t]
\caption{
Comparison of \methodname{} and its text-based counterpart on audio-to-image retrieval and visual question answering benchmarks.
}
\label{tab:image_ablation}
\centering
\resizebox{.9\textwidth}{!}{
\begin{tabular}{l ccc c c c c c}
\toprule
\multirow{2}{*}{Method} & \multicolumn{3}{c}{Input Mod.} & \multirow{2}{.1\textwidth}{\centering Pretrain Datasets} & Audio-to-Image Retrieval & \multicolumn{1}{c}{Visual QA (Acc.) $\uparrow$} & Latency $\downarrow$  \\
\cmidrule(lr){2-4} \cmidrule(lr){6-6} \cmidrule(lr){7-7} \cmidrule(lr){8-8}
& V & T & A & & \placesaudio{} (R@1 / R@5 / R@10) $\uparrow$ & \vqa{} & (ms)\\
\midrule
\methodname{}  & \checkmark & \checkmark & & - & 13.0 / 35.9 / 49.7 & 47.0 & 2010\\
\methodname{}  & \checkmark & & \checkmark & - & 12.7 / 33.3 / 48.0 & 46.7 & 52\\
\methodname{}  & \checkmark & \checkmark & & HT100M & 50.4 / 78.2 / 87.0 & 62.1 & 2010\\
\methodname{}  & \checkmark & & \checkmark & HT100M & 48.7 / 77.9 / 86.0 & 60.8 & 52\\
\methodname{}  & \checkmark & \checkmark & & YTT-S & \textbf{54.3} / \textbf{78.9} / \textbf{88.8} & \textbf{63.2} & 2010\\
\methodname{}  & \checkmark & & \checkmark & YTT-S & 49.0 / 78.2 / 86.8 & 61.0 & 52\\
\bottomrule
\end{tabular}
}
\end{table}

\subsection{Comparison to Text-based Counterpart}
\label{sec:comparison_to_text_counterpart}

\looseness=-1\Cref{tab:video_ablation} shows that \methodname{} outperforms the text-based counterpart in audio-to-video retrieval tasks when pretrained on either \howtohundredM{} or \Yttsubset{}.
On \cmumosei{} sentiment analysis, \methodname{} also outperforms its text variant when pretrained on \Yttsubset{}.
In \Cref{tab:image_ablation}, although \methodname{} slightly underperforms the text-based counterpart on audio-to-image retrieval and visual question answering, \methodname{} can still achieve decently comparable results and remain competitive while being 27x faster during inference due to the removal of ASR from the processing pipeline. More details on efficiency analysis are given in \cref{sec:efficiency}. The results provide evidence of the possibility of learning a more compact and efficient \vl{} representation from raw visual and audio signals compared to the prevailing VL learning paradigms with explicit text-based modules in the pipeline.

\begin{wraptable}[9]{r}{80mm}
\centering
\vspace{-35pt}
\caption{Latency of FFT, ASR and VL Models.}
\label{tab:efficiency}
\vspace{4pt}
\resizebox{0.58\textwidth}{!}{
\begin{tabular}{l c c cccc}
\toprule
\multirow{2}{*}{Model} & \multirow{2}{*}{\# Param} & Video Input & \multicolumn{4}{c}{Latency (ms) $\downarrow$} \\
\cmidrule(lr){3-3}
\cmidrule(lr){4-7}
& & Length / \# Frames & FFT & ASR & VL & Total \\
\midrule
\multirow{2}{*}{ASR-SpBr} & \multirow{2}{*}{195M} & 10s / 4 & - & 2110 & - & -\\
 & & 20s / 8 & - & 2890 & - & - \\
\midrule
\multirow{2}{*}{\methodname{}} & \multirow{2}{*}{88M} & 10s / 4 & 40 & - & 40 & 80\\
& & 20s / 8 & 60 &  - & 43 & 103 \\
\midrule
\multirow{2}{*}{\methodname{} + text} & 88M + 195M & 10s / 4 & - & 2110 & 25 & 2135 \\
& 88M + 195M & 20s / 8 & - & 2890 & 26 & 2916 \\
\midrule
\avlnet{} & 158M & 10s / 4 & 40 & - & 208 & 248 \\
\avlnet{} + text & 158M + 195M & 10s / 4 & - & 2110 & 206 & 2316 \\
\bottomrule
\end{tabular}
}
\end{wraptable}

\subsection{Efficiency Comparison}
\label{sec:efficiency}

To test inference latency, we sample 100 videos in \cmumosei{}.
As the average video length in the \cmumosei{} dataset is 12 seconds, we measure the latency with two sets of input video lengths: 10 and 20 seconds.
For 10s and 20s videos, we also use 4 and 8 video frames, respectively.
Then we calculate the processing time of Fast Fourier Transform (FFT), SpeechBrain (ASR-SpBr)~\cite{speechbrain}, TVLT, text-based TVLT, and AVLNet on the sampled inputs.
SpeechBrain is the default ASR module that we used in our text-based counterpart pipeline (see \cref{sec:ASR}).

As shown in \Cref{tab:efficiency}, we find that ASR dominates the inference time for text-based models.
Although ASR helps reduce the input length in transformers (as indicated by the VL module latency decrease), TVLT is more than 27x and 28x faster than text-based TVLT for inference with video input lengths of 10s and 20s, respectively, with only 1/3 of the parameters. The comparison is also shown in \cref{fig:teaser}.
In the bottom rows, we also show the inference latency of \avlnet{} and its text variant, where TVLT is 3x faster than \avlnet{} which contains audio-specific convolution modules.

\begin{wraptable}[8]{r}{78mm}
\centering
\vspace{-18pt}
\caption{Text vs. Speech Query for Video Retrieval.
}
\label{tab:tts_retrieval}
\vspace{5pt}
\resizebox{0.56\textwidth}{!}{
\begin{tabular}{l c c c}
\toprule
\multirow{2}{*}{Method} & \multirow{2}{.1\textwidth}{\centering Pretrain Datasets} & \multirow{2}{*}{Query} &  \multicolumn{1}{c}{Video Retrieval (R@1) $\uparrow$}  \\ \cmidrule(lr){4-4}
& & & \msrvtt{} \\
\midrule
\methodname{} & HT100M & Caption & 22.0\\
\methodname{} & HT100M & Speech Audio (TTS) & 20.1\\
\midrule
HERO \cite{li2020hero} &  HT100M & Caption & 16.8 \\
DeCEMBERT \cite{tang2021decembert} & HT100M, TVQA & Caption & 17.5 \\
ClipBERT \cite{lei2021Clipbert} &  COCO, VG & Caption & 22.0 \\
\avlnet{} \cite{rouditchenko2020avlnet} &  HT100M & Caption & 22.5 \\
\bottomrule
\end{tabular}
}
\end{wraptable}

\subsection{Text Query vs. Speech Query for Language-based Video Retrieval}
\label{Sec:TTS_text2video}

For text-to-video retrieval tasks, text captions are commonly used for queries~\cite{xu2016msr}. 
In \cref{sec:comparison_to_text_counterpart}, we show the experiment of audio-to-video retrieval tasks following \avlnet{}~\cite{rouditchenko2020avlnet}, where the audio queries are the sounds of the original videos.
Since video sounds and text captions have different information, the audio-to-video retrieval results are not directly comparable to the results in other text-to-video retrieval papers.
For a better comparison, we experiment with video retrieval based on two language queries: 1) text captions and 2) speech audio obtained by TTS (see \cref{sec:TTS}) from text captions.
\Cref{tab:tts_retrieval} shows \msrvtt{} video retrieval results of \methodname{} with text/audio queries and recent text-to-video retrieval models pretrained with a similar scale of data.\footnote{We exclude the models pretrained on large-scale image captions such as Conceptual Captions~\cite{sharma2018conceptual} that has written annotation, or visual encoder pretrained on a large-scale dataset beyond the scale of ImageNet~\cite{deng2009imagenet}, such as CLIP~\cite{radford2021learning}, as they are not directly comparable to our models.}
Although \methodname{} with audio
query slightly underperforms its text query counterpart due to TTS errors, it still outperforms other text-to-video retrieval models (HERO \cite{li2020hero} and DeCEMBERT \cite{tang2021decembert}), showing promising possibilities of speech-based video retrieval.

\begin{table}[h]
\caption{
\methodname{} on \cmumosei{} emotion analysis test set; \textit{WA}=weighted accuracy, \textit{F1}=weighted f1.
}
\label{tab:cmu_emo}
\centering
\resizebox{.8\textwidth}{!}{
\begin{tabular}{l ccc cccccc cccccc}
\toprule
\multirow{2}{*}{Method} & \multicolumn{3}{c}{Input Mod.} & \multicolumn{2}{c}{Happy} & \multicolumn{2}{c}{Sad} & \multicolumn{2}{c}{Angry} & \multicolumn{2}{c}{Fear} & \multicolumn{2}{c}{Disgust} & \multicolumn{2}{c}{Surprise} \\
\cmidrule(lr){2-4} \cmidrule(lr){5-6} \cmidrule(lr){7-8} \cmidrule(lr){9-10} \cmidrule(lr){11-12} \cmidrule(lr){13-14} \cmidrule(lr){15-16}
& V & T & A & WA & F1 & WA & F1 & WA & F1 & WA & F1 & WA & F1 & WA & F1 \\
\midrule
\methodname{}  & \checkmark & \checkmark & & 64.7 & 63.9 & 70.2 & 66.0 & 68.9 & 71.8 & 66.2 & 84.4 & \textbf{70.7} & \textbf{82.9} & 58.4 & 86.2 \\
\methodname{}  & \checkmark & & \checkmark & \textbf{65.1} & \textbf{64.1} & \textbf{72.2} & \textbf{70.0} & \textbf{69.9} & \textbf{72.1} & \textbf{68.1} & \textbf{88.0} & 68.8 & 79.6 & \textbf{62.1} & \textbf{87.4} \\
\bottomrule
\end{tabular}
}
\end{table}
\vspace{-5pt}

\subsection{Emotion Analysis}
Since TVLT takes raw visual and audio input instead of relying solely on text as in text-based TVLT, we further investigate what type of information TVLT can learn beyond speech on \cmumosei{} emotion classification task.
As shown in \Cref{tab:cmu_emo}, \methodname{} outperforms the text-based counterpart in most emotion categories, except for `Disgust'. 
We conjecture that TVLT is capable of capturing speech-related acoustic information, such as tone and loudness, which is helpful in recognizing these emotions, while this ability is absent from text-based ASR-dependent models.

\begin{table}[h]
\caption{
Finetuning performance on audio-to-video retrieval and multimodal sentiment analysis benchmarks.
For a fair comparison, we gray out the models that use ground-truth text transcription as additional input for \cmumosei{}.
}
\label{tab:video_comparison_results}
\centering
\resizebox{.9\textwidth}{!}{
\begin{tabular}{l ccc c ccc c }
\toprule
\multirow{2}{*}{Method} & \multicolumn{3}{c}{Input Mod.} & \multirow{2}{.1\textwidth}{\centering Pretrain Datasets} & \multicolumn{3}{c}{Audio-to-Video Retrieval (R@1) $\uparrow$} & Sentiment (A2) $\uparrow$   \\
\cmidrule(lr){2-4} \cmidrule(lr){6-8} \cmidrule(lr){9-9}
& V & T & A & & \msrvtt{} & \youcook{} & \crosstask{} & \cmumosei{} \\
\midrule
Multilogue-Net~\cite{shenoy-sardana-2020-multilogue}  & \checkmark &  & \checkmark & - & - & - & - & 75.2 \\
\avlnet{}~\cite{rouditchenko2020avlnet}      & \checkmark & & \checkmark & HT100M & 20.1 & 30.7 & 13.8 & -\\
\methodname{} (Ours)  & \checkmark & & \checkmark & HT100M & 22.6 & 31.8 & 14.9 & 75.3\\
\methodname{} (Ours) & \checkmark & & \checkmark & YTT-S & \textbf{23.8} & \textbf{32.8} & \textbf{15.3} & \textbf{76.8} \\
\bottomrule
\end{tabular}
}
\end{table}
\vspace{-5pt}

\begin{table}[h]
\caption{
Finetuning performance on audio-to-image retrieval and visual question answering (Visual QA).
For Visual QA, we create spoken questions from text via TTS (\cref{sec:TTS}).
$^\dagger$CSC (Conceptual Spoken Caption) is 3.3M image-speech pairs, where speech is obtained via TTS API from Conceptual Captions. The CSC dataset is not publicly available.
}
\label{tab:image_comparison_results}
\centering
\resizebox{.9\textwidth}{!}{
\begin{tabular}{l ccc c c c}
\toprule
\multirow{2}{*}{Method} & \multicolumn{3}{c}{Input Mod.} & \multirow{2}{.1\textwidth}{\centering Pretrain Datasets} & Audio-to-Image Retrieval & \multicolumn{1}{c}{Visual QA (Acc.) $\uparrow$}  \\
\cmidrule(lr){2-4} \cmidrule(lr){6-6} \cmidrule(lr){7-7}
& V & T & A & & \placesaudio{}  (R@1 / R@5 / R@10) $\uparrow$ & VQAv1 / \vqa{}\\
\midrule
TextMod~\cite{zhang2017speech} & \checkmark & \checkmark & & - & -  & 56.7 / - \\
\midrule
SpeechMod~\cite{zhang2017speech} & \checkmark & & \checkmark & - & -  & 47.0 / - \\
\avlnet{}~\cite{rouditchenko2020avlnet} & \checkmark & & \checkmark & HT100M & 44.8 / 76.9 / 86.4 & - \\
MILAN~\cite{Sanabria2021MILAN} & \checkmark & & \checkmark & CSC$^\dagger$ & \textbf{53.4} / \textbf{79.1} / 86.3 & - \\
\methodname{} (Ours) & \checkmark & & \checkmark & HT100M & 48.7 / 77.9 / 86.0 & 58.6 / 60.8\\
\methodname{} (Ours) & \checkmark & & \checkmark & YTT-S & 49.0 / 78.2 / \textbf{86.8} & \textbf{58.9} / \textbf{61.0} \\
\bottomrule
\end{tabular}
}
\end{table}
\vspace{-5pt}

\subsection{Comparison to State-of-the-art Textless Models}
\label{sec:results_comparison_to_sota}

We compare our \methodname{} with recent models that also take raw visual and audio signals as input but involve audio-specific designs in their networks. As shown in \Cref{tab:video_comparison_results}, \methodname{} outperforms \avlnet{}~\cite{rouditchenko2020avlnet} on three audio-to-video retrieval (\msrvtt{}, \youcook{}, \crosstask{}) tasks and outperform Multilogue-Net~\cite{shenoy-sardana-2020-multilogue} on multimodal sentiment analysis (\cmumosei{}) task with a simple modality-agnostic design. Similarly, \Cref{tab:image_comparison_results} shows that \methodname{} achieves competitive results with \avlnet{}~\cite{rouditchenko2020avlnet} and MILAN~\cite{Sanabria2021MILAN} on audio-to-image retrieval (\placesaudio{}). Note that MILAN\footnote{The dataset is also not publicly available.} is pretrained on Conceptual Spoken Caption~\cite{Ilharco2019} which contains 3.3M well-aligned image-speech pairs taken from Conceptual Captions~\cite{sharma2018conceptual} with TTS generated speech, whereas our TVLT is able to elicit effective representation from video inputs where \vl{} clues are only weakly aligned.
TVLT also outperforms both variants of the VQA models (TextMod, SpeechMod) in \citet{zhang2017speech} on \vqaold{}.

\subsection{Ablation Studies}
\label{sec:ablation}
In the following, we show the results of the ablation study on \methodname{} training details: the audio masking strategy, the encoder/decoder architectures, and the pretraining objectives.

\begin{wraptable}[9]{r}{65mm}
\centering
\vspace{-5pt}
\caption{Audio masking configurations.
}
\label{tab:ablation_audio_masking}
\vspace{6pt}
\centering
\resizebox{0.45\textwidth}{!}{
\begin{tabular}{cc cc}
\toprule
\multirow{2}{*}{Patch Size} & \multirow{2}{.1\textwidth}{\centering Masking on speech} & \msrvtt{} & \vqa{} \\
& & (R@1) & (Acc.)\\
\midrule
$16\times 16$ &          & 21.7 & 57.8 \\
$16\times 16$ & \checkmark & \textbf{22.3} &   58.6 \\
$2\times 128$ &       &  21.0 & 58.8 \\
$2\times 128$ & \checkmark &  21.2 &   \textbf{59.2}   \\
\bottomrule
\end{tabular}
}
\end{wraptable}

\noindent\textbf{Audio Masking Strategy.}
In \Cref{tab:ablation_audio_masking}, we show the result of finetuning performance with different audio masking configurations, described in \cref{sec:masking_strategy}.
For patch sizes, masking audio patches on detected speech spans improves performance across the board.
However, we did not observe strict superiority between the two patch sizes;
$2\times 128$ achieves higher scores on \msrvtt{}, while $16\times 16$ achieves higher scores on \vqa{}.
For our default pretraining configuration,
we use the $16\times 16$ patch size and use speech span detection, since the $16\times 16$ sized patch is also used in visual embedding (thus modality-agnostic) and speech span detection improves performance with minimal additional computation (see appendix).

\begin{wraptable}[7]{r}{50mm}
\centering
\vspace{-15pt}
\caption{Encoder variants.}
\label{tab:ablation_encoder_architecture}
\vspace{6pt}
\centering
\resizebox{0.35\textwidth}{!}{
\begin{tabular}{l cc}
\toprule
\multirow{2}{*}{Encoder} &  \msrvtt{} & \vqa{}\\
 & (R@1) &  (Acc.)\\
\midrule
Separate &   9.6 & 53.1  \\
Joint    &    \textbf{10.2}   &   \textbf{54.6}  \\
\bottomrule
\end{tabular}
}
\end{wraptable}

\noindent\textbf{Encoder Architecture.}
As described in \Cref{sec:enc_dec}, we use the joint encoder in \methodname{}.
We compare this to modality-specific encoders for vision and audio.
\Cref{tab:ablation_encoder_architecture} below compares the separate encoders with the joint encoder for two tasks: \vqa{} and \msrvtt{}.
To tackle \vqa{} with separate encoders, we learned a two-layer self-attention fusion layer over the concatenation of hidden states of the vision and audio encoder. Our joint encoder architecture achieves better accuracy on both tasks than a separate encoder architecture.
The results show that although vision and audio spectrogram are two different modalities, the single joint encoder could learn useful cross-modal representation for VL tasks without needing modality-specific encoders.

\begin{wraptable}[8]{r}{50mm}
\centering
\vspace{-15pt}
\caption{Decoder variants.}
\label{tab:ablation_decoder_architecture}
\vspace{6pt}
\centering
\resizebox{0.35\textwidth}{!}{
\begin{tabular}{l cc}
\toprule
\multirow{2}{*}{Decoder} &  \msrvtt{} & \vqa{}\\
 & (R@1) &  (Acc.)\\
\midrule
Separate &   \textbf{22.3}    &   \textbf{58.6}   \\
Joint    &    22.0   &   58.1  \\
\bottomrule
\end{tabular}
}
\end{wraptable}

\noindent\textbf{Decoder Architecture.}
As described in \cref{sec:masked_autoencoding}, we use separate decoders (with shared weights) for the vision and audio MAE pretraining objectives.
We compare this separate decoding with joint decoder, where we feed the concatenated encoder outputs to the decoder and jointly reconstruct the video frames and spectrogram.
\Cref{tab:ablation_decoder_architecture} shows that pretraining with separate decoder outperforms joint decoder on finetuning performance, while being more efficient as well.

\begin{wraptable}[8]{r}{50mm}
\centering
\vspace{-20pt}
\caption{Pretraining objectives.}
\label{tab:ablation_objective}
\vspace{3pt}
\centering
\resizebox{0.35\textwidth}{!}{
\begin{tabular}{l cc}
\toprule
\multirow{2}{*}{Objectives} &  \msrvtt{} & \vqa{}\\
 & (R@1) &  (Acc.)\\
\midrule
Random init & 4.3         & 46.7 \\
VAM &        21.0       & 56.2 \\
MAE &        18.6       &  54.1 \\
VAM + MAE &       \textbf{22.3}        & \textbf{58.6}\\
\bottomrule
\end{tabular}
}
\end{wraptable}

\noindent\textbf{Pretraining Objectives.}
We measure the impact of each pretraining objective described in \cref{sec:pretrain_objectives}.
\Cref{tab:ablation_objective} shows that each of the pretraining objectives (MAE and VAM) improves finetuning performance over random weight initialization. The combination of VAM and MAE further improves the finetuning performance, and we use this configuration as default for \methodname{} pretraining.

\section{Conclusion}
\vspace{-5pt}
\label{sec:conclusion}
In this work, we present \methodname{}, a simple end-to-end \vl{} transformer that can accept low-level visual and audio signals for \vl{} representation learning.
Our \methodname{} achieves competitive performance with other state-of-the-art audio-based \vl{} models on
visual question answering,
image retrieval, video retrieval,
and multimodal sentiment analysis.
We also show that by eliminating the need for expensive ASR in the model pipeline, \methodname{} can be 28x faster than its text-based counterpart while achieving comparable performance.
We comprehensively analyze the efficiency of our model and show ablation studies over different training variants.
We hope that our research will inspire further exploration of simple and efficient \vl{} frameworks with low-level signals.

\section*{Acknowledgments}
We thank the reviewers for their helpful comments.
This work was supported by ARO  Award W911NF2110220, DARPA KAIROS Grant FA8750-19-2-1004, ONR Grant N000141812871, and NSF-AI Engage Institute DRL-211263. The views, opinions, and/or findings contained in this article are those of the authors and not of the funding agency.

{\small
\bibliographystyle{acl_natbib}
\bibliography{references}
}

\section*{Checklist}


\begin{enumerate}
\item For all authors...
\begin{enumerate}
  \item Do the main claims made in the abstract and introduction accurately reflect the paper's contributions and scope?
    \answerYes{}
  \item Did you describe the limitations of your work?
    \answerYes{See supplementary material}
  \item Did you discuss any potential negative societal impacts of your work?
    \answerYes{See supplementary material}
  \item Have you read the ethics review guidelines and ensured that your paper conforms to them?
    \answerYes{}
\end{enumerate}

\item If you are including theoretical results...
\begin{enumerate}
  \item Did you state the full set of assumptions of all theoretical results?
    \answerNA{}
        \item Did you include complete proofs of all theoretical results?
    \answerNA{}
\end{enumerate}

\item If you ran experiments...
\begin{enumerate}
  \item Did you include the code, data, and instructions needed to reproduce the main experimental results (either in the supplemental material or as a URL)?
    \answerYes{See supplemental material}
  \item Did you specify all the training details (e.g., data splits, hyperparameters, how they were chosen)?
    \answerYes{See \cref{sec:expsetup}}
        \item Did you report error bars (e.g., with respect to the random seed after running experiments multiple times)?
    \answerNo{}
        \item Did you include the total amount of compute and the type of resources used (e.g., type of GPUs, internal cluster, or cloud provider)?
    \answerYes{See \cref{sec:training_details}}
\end{enumerate}

\item If you are using existing assets (e.g., code, data, models) or curating/releasing new assets...
\begin{enumerate}
  \item If your work uses existing assets, did you cite the creators?
    \answerYes{}
  \item Did you mention the license of the assets?
    \answerYes{See supplementary material}
  \item Did you include any new assets either in the supplemental material or as a URL?
    \answerYes{}
  \item Did you discuss whether and how consent was obtained from people whose data you're using/curating?
    \answerNA{}
  \item Did you discuss whether the data you are using/curating contains personally identifiable information or offensive content?
    \answerNA{}
\end{enumerate}

\item If you used crowdsourcing or conducted research with human subjects...
\begin{enumerate}
  \item Did you include the full text of instructions given to participants and screenshots, if applicable?
    \answerNA{}
  \item Did you describe any potential participant risks, with links to Institutional Review Board (IRB) approvals, if applicable?
    \answerNA{}
  \item Did you include the estimated hourly wage paid to participants and the total amount spent on participant compensation?
    \answerNA{}
\end{enumerate}

\end{enumerate}


\newpage
\appendix

In this appendix, we include
the pretraining dataset combination experiment (\cref{Sec:pretrain_data_combination}),
TTS-based text-to-video retrieval experiment (\cref{Sec:TTS_text2video}),
ASR quality experiment (\cref{Sec:ASR_quality}),
implementation details (\cref{Sec:implementation_details}),
finetuning on unimodal ASR task (\cref{Sec:finetuning_ASR}),
visualization of MAE reconstruction (\cref{Sec:mae_vis}),
limitations and potential negative impacts (\cref{Sec:limitation_impacts}),
and licenses (\cref{Sec:license}).

\section{Combination of Pretraining Datasets}
\label{Sec:pretrain_data_combination}

\Cref{tab:video_ablation} and \Cref{tab:image_ablation} in the main paper show that \methodname{} either pretraining on \howtohundredM{}~\cite{miech2019howto100m} or \Yttsubset{}~\cite{zellers2021merlot} can outperform random initialization across the board.
Among the two pretraining datasets,
models pretrained on \Yttsubset{} achieve higher performance than models pretrained on \howtohundredM{}.
The relative improvement is consistent with the findings of \citet{zellers2021merlot}, and we suspect that coverage of a wider range of video topics improves overall performance.
We also experiment with pretraining \methodname{} with the combination of \howtohundredM{} and \Yttsubset{}.
The total size of the pretraining dataset size is 1.85M = (0.92M  + 0.93M) videos, and we pretrain the model for 200k steps.
As shown in \Cref{tab:pretrain_comb}, pretraining on the combination of both datasets achieves better finetuning performance than
single-dataset pretraining on both \msrvtt{} audio-to-video retrieval and VQAv2. The results indicate that \methodname{} can take advantage of the domain diversity of \Yttsubset{} and that pretraining with data from a diverse range of domains can result in a more adaptable representation.

\begin{table}[h]
\centering
\caption{Finetuning performance of \methodname{} pretrained on different datasets.}
\label{tab:pretrain_comb}
\resizebox{0.99\textwidth}{!}{
\begin{tabular}{l ccc c cc}
\toprule
\multirow{2}{*}{Method} & \multicolumn{3}{c}{Input Mod.} & \multirow{2}{.1\textwidth}{\centering Pretrain Datasets} & \multicolumn{1}{c}{Audio-to-Video Retrieval (R@1) $\uparrow$} & \multicolumn{1}{c}{Visual QA (Acc.) $\uparrow$}\\
\cmidrule(lr){2-4} \cmidrule(lr){6-6} \cmidrule(lr){7-7}
& V & T & A & & \msrvtt{} & \vqa{} \\
\midrule
\methodname{} & \checkmark & & \checkmark  & \howtohundredM{} & 22.6 & 60.8 \\
\methodname{} & \checkmark & & \checkmark  & \Yttsubset{} & 23.8 & 61.0 \\
\methodname{} & \checkmark & & \checkmark  & \howtohundredM{}+\Yttsubset{} & \textbf{25.0} & \textbf{61.4}\\
\bottomrule
\end{tabular}
}
\end{table}

\begin{wraptable}{r}{70mm}
\centering
\vspace{-5pt}
\caption{\methodname{} with audio/text on \cmumosei{}.}
\label{tab:cmumosei_asr}
\vspace{8pt}
\resizebox{0.5\textwidth}{!}{
\begin{tabular}{l cc}
\toprule
\multirow{2}{*}{Language Input} & \multicolumn{2}{c}{\cmumosei{} (A2) $\uparrow$}  \\
\cmidrule(lr){2-3}
& HT100M & YTT-S \\
\midrule
Audio & 75.3 & \textbf{76.8} \\
Text (ASR-SpeechBrain)  & \textbf{76.5} & 76.6 \\
\midrule
Text (ASR-Google)  & 77.1 & 77.8  \\
Text (GT Transcripts) &  78.9 & 79.1\\
\bottomrule
\end{tabular}
}
\end{wraptable}

\section{Impact of ASR quality}
\label{Sec:ASR_quality}

\Cref{tab:cmumosei_asr} shows the results of \methodname{} on \cmumosei{} sentiment analysis with the following different inputs: audio, ASR-based text, and ground-truth text transcriptions.
ASR-Google and ASR-SpeechBrain refer to Google Cloud API and SpeechBrain, respectively (see main paper \cref{sec:ASR}).
Although \methodname{} pretrained on \howtohundredM{} underperform the text variant with SpeechBrain ASR input,
\methodname{} pretrained on \Yttsubset{} (76.8) achieves comparable results to those of the text variant with SpeechBrain ASR (76.6), which sheds light on the effectiveness of \methodname{}.
Although there is still a gap between \methodname{} and text-based \methodname{} with higher quality ASR or ground truth transcript input,
we expect that \methodname{} can be further improved with larger-scale pretraining (e.g., full \Yttemp{} dataset) on raw video signals.

To better understand the impact of ASR on downstream tasks, we show two examples of the \cmumosei{} sentiment analysis task in \Cref{tab:cmumosei_examples}.
For example (a), ASR-Google Cloud provides more accurate transcription than ASR-SpeechBrain, resulting in more accurate sentiment estimation (ASR-SpeechBrain: -1.0 vs. ASR-Google: 0.0; label: 0.0).
For example (b), ASR-Google Cloud and ASR-SpeechBrain provide similar transcription quality, resulting in the same sentiment estimation (ASR-SpeechBrain: 2.0 vs ASR-Google: 2.0; label: 1.0).

\begin{table}[h]
\caption{
Comparison of different ASR models on \cmumosei{} sentiment analysis task.
Sentiment label has range $[-3, 3]$, where -3 and 3 corresponds to negative and positive, respectively.
We use \methodname{} pretrained on \howtohundredM{}.
}
\label{tab:cmumosei_examples}
\centering
\resizebox{0.999\textwidth}{!}{
\begin{tabular}{c p{0.23\textwidth}p{0.23\textwidth}p{0.23\textwidth}P{0.1\textwidth}P{0.1\textwidth}P{0.1\textwidth}P{0.1\textwidth}}
\toprule
& GT Transcripts & ASR-SpeechBrain & ASR-Google Cloud & Pred (GT Transcripts) & Pred (ASR-SpBr) & Pred (ASR-GC) & Label\\
\midrule
(a) & This is a new movie (uhh) in which a character is confined to his house, he is under house arrest, and his mother takes away his Xboxes and TV as sort of a little bit of additional punishment & communicate of additional punishment thoroughly & Serbia this is a new movie and which a character is confined to his house. He is under house arrest and his mother takes away his Xboxes and TVs is sort of a little bit of additional punishment. & 0.0 & -1.0 & 0.0 & 0.0 \\
\midrule
(b) & The club that I'm part of that organize that has about currently 40 some students and then last year we had 260-something come out to the dance. & well the club that i'm part of that organizes it has about currently forty some students and then flashed here we had two hundred and sixteen something come out to the dance & The club that I'm part of that organizes it has about currently 40 some students. And then last year we had 260 something come out to the dance & 1.0 & 2.0 & 2.0 & 1.0 \\
\bottomrule
\end{tabular}}
\end{table}

\section{Implementation Details}
\label{Sec:implementation_details}

\subsection{Speech Span Detection}
\label{Sec:audio_span_detection_detail}

For the speech span detection mentioned in the main paper \cref{sec:masking_strategy}, we use the Audiotok~\cite{Sehili_auditok_2021} word-level speech event detector.
We use the configurations as follows:
(1) We set a single speech event to have a duration within [0.3s, 1.2s], so that an event is likely to cover a single word.
(2) We set $\texttt{max\_silence}=0.05s$. $\texttt{max\_silence}$ refers to the maximum silence gap between two speech spans. If the silence gap is too large, it is usually a stop between two words. Therefore, setting a low value ensures that we do not detect two words as a single word.
(3) We use an energy threshold of 70, which is higher than the default value of 55, to avoid false positives of detecting noise. This is because real-world audio contains natural sounds and noises that usually come with a high level of audio signal energy.
In the speech spans detected on \howtohundredM{}, each word has an average length of 15 in our audio spectrogram (\cref{sec:embeddings}).
As this is similar to the size of a single audio patch (16x16), masking an audio patch usually covers a word in speech.

\begin{table}[h]
\caption{Audio Pipeline Latency.}
\label{tab:audio_pipeline_latency}
\centering
\resizebox{0.99\textwidth}{!}{
\begin{tabular}{c ccc cc}
\toprule
\multirow{2}{.1\textwidth}{\centering Audio Length} & \multicolumn{3}{c}{CPU Latency (ms) $\downarrow$} & \multicolumn{2}{c}{GPU Latency (ms) $\downarrow$} \\
\cmidrule(lr){2-4} \cmidrule(lr){5-6}
 & Data Loading & Fast Fourier Transform & Speech Span Detection & ASR & VL Model\\
\midrule
10s & 60 & 40 & 130 & 2110 & 40\\
20s & 110 & 60 & 170 & 2890 & 43\\
\bottomrule
\end{tabular}
}
\end{table}

\subsection{Audio Pipeline latency}
\label{Sec:audio_pipeline_latency}

In \Cref{tab:audio_pipeline_latency}, we show the detailed latency for each audio processing pipeline for two different audio length settings: 10s and 20s.
In both settings, ASR takes significantly longer processing time than all other modules and becomes the bottleneck of the entire \vl{} pipeline.

\section{Finetuning on Unimodal ASR Task}
\label{Sec:finetuning_ASR}
\begin{wraptable}{r}{50mm}
\centering
\vspace{-20pt}
\caption{Finetuning on ASR.}
\label{tab:finetuning_asr}
\vspace{6pt}
\centering
\resizebox{0.35\textwidth}{!}{
\begin{tabular}{lcc}
\toprule
\multirow{2}{*}{Encoder PT} & \multicolumn{2}{c}{WER (\%) $\downarrow$} \\
\cmidrule{2-3}
& dev-clean & dev-other\\
\midrule
No-pretrain	& 3.1 & 6.0 \\
V+A pretrain & \textbf{2.3} & \textbf{4.7} \\
\bottomrule
\end{tabular}
}
\end{wraptable}

To explore whether the cross-modal representation of \methodname{} is useful for unimodal tasks, we experiment using \methodname{} as an audio encoder for an ASR model. Specifically, we construct a 4-layer transformer language model that attends to \methodname{} encoder outputs via cross-attentions and jointly train the encoder and decoder. We experiment with two settings: where the \methodname{} encoder is randomly initialized or initialized with V+A pretraining. We train the models on LibriSpeech~\cite{panayotov2015librispeech}, a widely used ASR corpus with 960 hours of English audiobooks, and evaluate them on its two dev sets, dev-clean and dev-other. As shown in \Cref{tab:finetuning_asr}, our ASR model with V+A pretrained \methodname{} encoder outperforms the No-pretrain baseline by 0.8 (dev-clean) and 1.3 (dev-other) in Word Error Rate (WER), respectively. The results show that the cross-modal representation learned by \methodname{} could also be helpful for ASR, a unimodal task.

\section{MAE Reconstruction Visualization}
\label{Sec:mae_vis}
\begin{figure*}[t]
  \centering
  \includegraphics[width=0.32\textwidth]{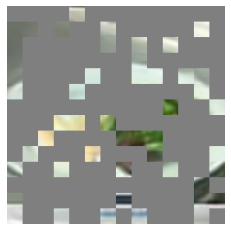}
  \includegraphics[width=0.32\textwidth]{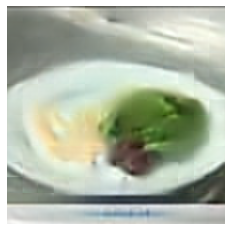}
  \includegraphics[width=0.32\textwidth]{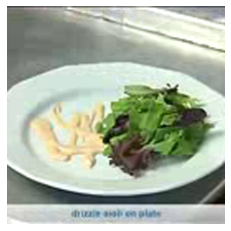}
  \includegraphics[width=0.32\textwidth]{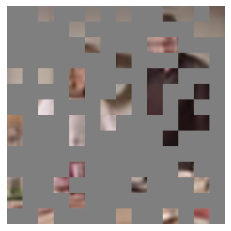}
  \includegraphics[width=0.32\textwidth]{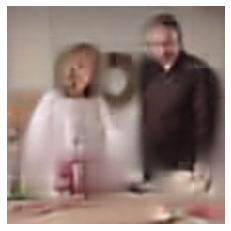}
  \includegraphics[width=0.32\textwidth]{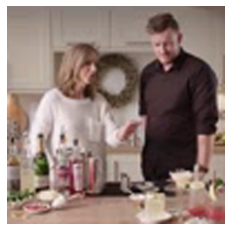}
  \includegraphics[width=0.32\textwidth]{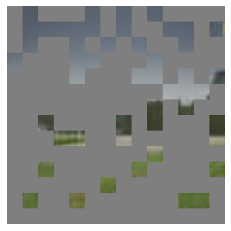}
  \includegraphics[width=0.32\textwidth]{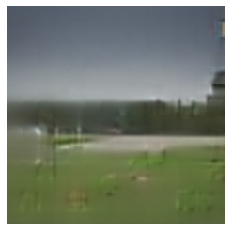}
  \includegraphics[width=0.32\textwidth]{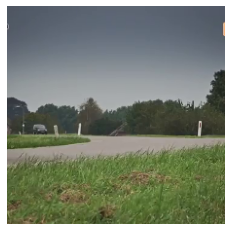}
  \caption{
    Visualization on video frames reconstruction (single frame): masked frames (left), reconstruction (middle), and original frames (right).
}
\label{fig:mae_vis_video}
\end{figure*}

\begin{figure*}[t]
  \centering
  \includegraphics[width=0.32\textwidth]{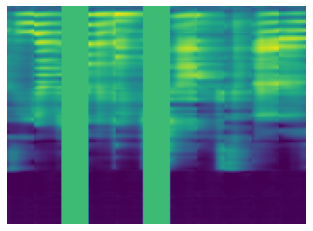}
  \includegraphics[width=0.32\textwidth]{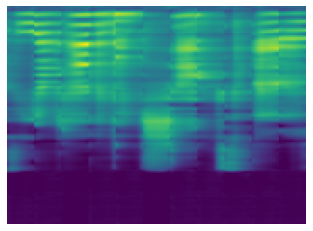}
  \includegraphics[width=0.32\textwidth]{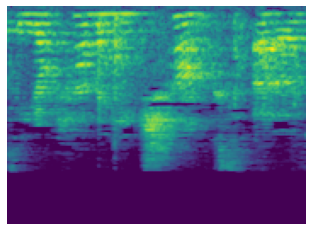}
  \includegraphics[width=0.32\textwidth]{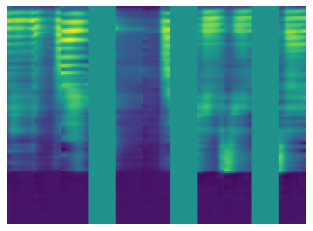}
  \includegraphics[width=0.32\textwidth]{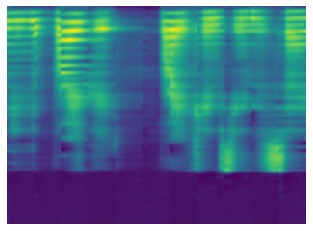}
  \includegraphics[width=0.32\textwidth]{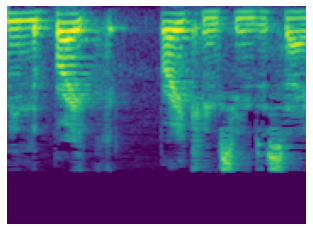}
  \includegraphics[width=0.32\textwidth]{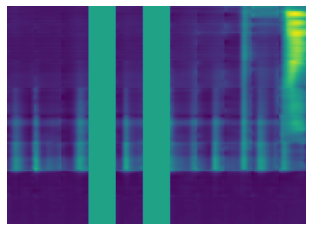}
  \includegraphics[width=0.32\textwidth]{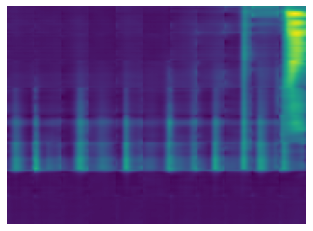}
  \includegraphics[width=0.32\textwidth]{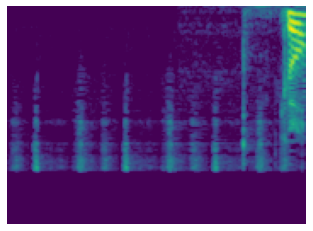}
  \caption{
    Visualization on video frames reconstruction: masked audio spectrogram (left), reconstruction (middle), and original audio spectrogram (right).
}
\label{fig:mae_vis_audio}
\end{figure*}

In \Cref{fig:mae_vis_video} and \Cref{fig:mae_vis_audio}, we show the reconstruction results with the MAE head, described in the main paper \cref{sec:masked_autoencoding}.
In each figure,
the left column shows the masked input,
the middle column shows the reconstruction, 
and the right column shows the target.
We use masking ratio 0.75, image size $224 \times 224$, and audio spectrogram size 176 $\times$ 128 (time $\times$ frequency) for this visualization.

\section{Limitations}
\label{Sec:limitation_impacts}

\paragraph{Green AI.}
A key barrier to the adoption of Green AI~\cite{schwartz2020green} has been the incentive to use massive computational power for pretraining.
As shown in our main paper, \methodname{} is also subject to pretraining in order to achieve decent performance on visual linguistic tasks. While \methodname{} is substantially faster than \vl{} models with explicit text-based modules that can help reduce pretraining computation, there is still scope for future work on energy-efficient training to alleviate the heavy reliance on large-scale pretraining. 

\paragraph{English-only Datasets.}
We perform transfer learning with \methodname{} pretrained with \howtohundredM{} and \Yttemp{} datasets. Both datasets mostly contain content in English, since \howtohundredM{}~\cite{miech2019howto100m} videos are obtained from English queries, and the authors of \Yttemp{}~\cite{zellers2021merlot} filtered out videos with non-English ASR results.
Therefore, our models pretrained with the two datasets might not have a good performance on non-English tasks without additional pretraining.

Note that the \methodname{} framework is a language-agnostic method, so one can adapt our model to a non-English dataset without any architectural change.
Furthermore, our architecture eliminates the need for external ASR modules, which reduces the computation of the typical \vl{} pipeline.
To reduce environmental damage, we will publicly release our code and pretrained checkpoint.

\section{License}
\label{Sec:license}

We will publicly release our code and models. We use standard licenses from the community and provide the following links to the licenses for the datasets, codes, and models that we used in the project. For more details, see the individual link.
\vspace{3pt}

\noindent\textbf{\howtohundredM{}:} \href{https://github.com/antoine77340/howto100m/blob/master/LICENSE}{Apache}

\noindent\textbf{\Yttemp{}:} 
\href{https://github.com/rowanz/merlot/blob/main/LICENSE}{MIT}

\vspace{3pt}
\noindent\textbf{PyTorch:} \href{https://github.com/pytorch/pytorch/blob/master/LICENSE}{BSD-style}

\vspace{3pt}
\noindent\textbf{Huggingface Transformers:} \href{https://github.com/huggingface/transformers/blob/master/LICENSE}{Apache}

\vspace{3pt}
\noindent\textbf{Torchvision:} \href{https://github.com/pytorch/vision/blob/master/LICENSE}{BSD 3-Clause}

\end{document}